\documentclass[letterpaper,10pt,conference]{ieeeconf}

\IEEEoverridecommandlockouts
\usepackage{graphicx}
\usepackage{booktabs}
\usepackage{multirow}
\usepackage{amsmath}
\usepackage{amssymb}
\usepackage{caption}

\newenvironment{packed_item}{
\begin{itemize}
  \setlength{\itemsep}{1pt}
  \setlength{\parskip}{2pt}
  \setlength{\parsep}{0pt}
}{\end{itemize}}

\newcommand{\shortname}{DexSeed}
\newcommand{\grabseqnum}{19}
\newcommand{\graspmodenum}{8}
\newcommand{\vlaseqnum}{25270}

\usepackage[hidelinks]{hyperref}
\usepackage[capitalize]{cleveref}
\hypersetup{
  pdftitle={DexSeed: Expanding Human Hand--Object Interactions into Robot Demonstrations for Dexterous Policy Learning},
  pdfauthor={Huaijin Pi, Zehong Shen, Yong Wang, Kechun Xu, Changpeng Wang, Dingkun Wei, Xueyang Wang, Wenjia Wang, Xing Zhu, Yujun Shen, Taku Komura}
}

\title{\LARGE \bf
\shortname{}: Expanding Human Hand--Object Interactions into Robot Demonstrations for Dexterous Policy Learning
}

\author{%
Huaijin Pi$^{1,2}$,
Zehong Shen$^{2,*}$,
Yong Wang$^{3}$,
Kechun Xu$^{2}$,
Changpeng Wang$^{3}$\\
Dingkun Wei$^{2,3}$,
Xueyang Wang$^{2,3}$,
Wenjia Wang$^{1}$,
Xing Zhu$^{2}$,
Yujun Shen$^{2,\dagger}$,
Taku Komura$^{1,\dagger}$\\[3pt]
{\small $^{*}$Project Lead \qquad $^{\dagger}$Corresponding Authors}\\[3pt]
{\small $^{1}$The University of Hong Kong \qquad $^{2}$Robbyant \qquad $^{3}$Zhejiang University}%
\thanks{Huaijin Pi, Yong Wang, Changpeng Wang, Dingkun Wei, and Xueyang Wang conducted this work during their internships at Ant Group.
This work was supported by Ant Group Research Intern Program.}%
}

\begin{document}
\maketitle
\thispagestyle{plain}
\pagestyle{plain}

\begin{abstract}
We study how to generate diverse, physically executable robot demonstrations for dexterous vision-language-action (VLA) learning.
Existing demonstration-generation methods typically rely on costly teleoperated examples.
Meanwhile, human hand--object datasets provide a reusable source, but existing transfer methods mainly reproduce given interactions.
Turning these interactions into VLA training data further requires coverage of different objects and manipulation locations, together with executable arm and finger actions, and visual observations.
To achieve this, we present \shortname{}, a framework for low-cost generation of dexterous robot data from human hand--object interactions.
Our key idea is to expand human sequences while preserving their grasps and realizing the resulting motions under arm-tracking dynamics in physics simulation.
Specifically, we organize source sequences by grasp mode with shared language descriptions.
InteractionMesh-based retargeting and arm trajectory augmentation adapt these sequences to the robot hand, different object geometries, and pick-and-place locations while preserving local hand--object relationships.
An RL finger-control policy produces executable rollouts, while the arm tracks precomputed trajectories.
Rendered observations and recorded actions from successful rollouts are paired with grasp descriptions to fine-tune a VLA.
Experiments show improved contact fidelity in retargeting and more reliable whole-arm execution than the evaluated baselines.
Simulation evaluation and real-world deployment further demonstrate that a VLA fine-tuned solely on the generated demonstrations can execute language-specified grasps.
Project page: \url{https://robbyant-research.github.io/DexSeed/}.
\end{abstract}

\section{Introduction}
\label{sec:intro}
Dexterous hands enable robots to manipulate objects through coordinated finger motion and rich hand--object contacts.
Vision-language-action (VLA) models learn to map visual observations and language instructions to robot actions, providing a framework for learning manipulation skills from demonstrations~\cite{24arxiv_pi0}.
Applying these models to dexterous manipulation requires diverse robot data that capture coordinated arm and finger actions together with visual observations and language instructions.
In this work, we study how to generate physically executable robot demonstrations at scale to support dexterous VLA learning.

Existing demonstration-generation methods expand robot datasets across object poses, scene layouts, and motion trajectories, but typically rely on teleoperated demonstrations as their starting point~\cite{23corl_mimicgen,25icra_dexmimicgen,25rss_demogen}.
Collecting these source demonstrations is costly for dexterous hands, as teleoperation requires coordinating many joints while maintaining stable hand--object contact~\cite{20icra_dexpilot,23rss_anyteleop,24corl_opentelevision}.
Reusing existing human hand--object datasets offers an alternative source of diverse manipulation behaviors without collecting new teleoperated demonstrations~\cite{20eccv_grab,22cvpr_oakink}.
Retargeting and physics-based imitation methods transfer these interactions to robots by addressing differences in hand morphology and contact dynamics~\cite{25arxiv_spider,25cvpr_maniptrans,25iclr_dextrack,26arxiv_toporetarget,26arxiv_c2dex,26arxiv_regrind,26arxiv_doasido}.
Expanding these transferred interactions into VLA training datasets requires varying object geometries and manipulation locations while retaining the demonstrated grasps.
Such variations can alter hand--object contacts, and arm-tracking errors during execution can further disrupt the intended interaction.
The resulting demonstrations must therefore preserve the grasp across these variations and provide physically executable arm and finger actions paired with visual observations.

\begin{figure}[t]
    \centering
    \includegraphics[width=1.0\columnwidth]{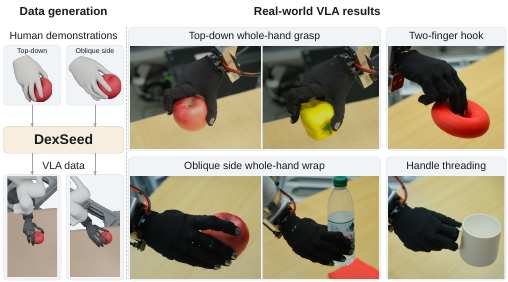}
    \caption{\textbf{From human data to robot execution.}
    DexSeed organizes human interactions by grasp mode and expands them into simulated robot data.
    The VLA trained on these data executes language-specified grasps on real objects.
    The examples show two grasp modes applied to the same apple and the same grasp mode applied to different objects.
    }
    \label{fig:teaser}
\end{figure}

To address this need, we propose \shortname{}, a framework for low-cost generation of dexterous robot demonstrations from human interactions.
Our key idea is to organize human interactions by grasp, expand them while preserving hand--object relationships, and obtain executable demonstrations through reinforcement learning with the robot arm included in simulation.
Specifically, we group human sequences by grasp similarity and select representative modes, each paired with a shared description of how to grasp.
InteractionMesh-based retargeting~\cite{12humanoids_interactionmesh} first adapts the sequence to the robot hand and compatible object geometries while preserving local hand--object relationships.
We combine the resulting hand references with whole-arm pick-and-place trajectories for different object starting positions and placement targets.
To realize these references under arm-tracking dynamics, we train an RL finger-control policy in IsaacLab~\cite{isaaclab} while the robot arm tracks precomputed joint trajectories, allowing the policy to learn finger corrections under the resulting wrist-tracking errors.
We render observations from the policy's successful rollouts and pair them with the recorded robot actions and corresponding grasp descriptions to form training demonstrations.
The resulting data are used to fine-tune a VLA that jointly controls the arm and fingers to execute language-specified grasps.

We select \graspmodenum{} representative grasp modes from the GRAB dataset \cite{20eccv_grab} and use \grabseqnum{} source sequences from these modes to generate a robot-centric dataset of \vlaseqnum{} demonstrations.
Compared with baselines, our retargeting method better preserves the intended hand--object interactions, while our physics-based imitation achieves more reliable execution of augmented references with variations in object geometry and trajectory.
We further show that a dexterous VLA fine-tuned exclusively on our simulation-generated demonstrations with rendered observations can be deployed on a real robot without collecting real-robot action demonstrations.

The contributions of this work are as follows.
\begin{packed_item}
    \item We present a framework that groups human interactions by grasp mode and expands individual sequences into diverse robot demonstrations paired with language descriptions of how to grasp.
    \item We develop an InteractionMesh-based data generation method that preserves hand--object interactions during retargeting and transfer across different object geometries. For each source sequence, a physics-based imitation policy is jointly trained on its augmented references to obtain executable robot rollouts.
    \item We construct a robot-centric dataset of \vlaseqnum{} demonstrations from \grabseqnum{} human sequences spanning \graspmodenum{} representative grasp modes, and demonstrate its utility through simulation evaluation and real-world deployment of a VLA fine-tuned solely on the generated data, without real-robot action demonstrations at deployment.
\end{packed_item}

\section{Related Work}
\label{sec:related}
\subsection{Dexterous Manipulation Data}
Dexterous robot learning relies on diverse demonstrations that capture high-dimensional hand actions, contact-rich object motion, and robot-centric observations.
Existing data sources include human hand--object datasets~\cite{20eccv_grab,22cvpr_oakink,24cvpr_oakink2}, direct robot teleoperation~\cite{20icra_dexpilot,23rss_anyteleop,24corl_opentelevision}, and systems for capturing interactions or generating synthetic demonstrations~\cite{24arxiv_dexcap,25arxiv_dexumi,25rss_dex1b}.
Human datasets contain diverse grasping and manipulation behaviors, while demonstrations collected directly on robots provide robot actions and observations but are costly to collect at scale.
Our work turns human hand--object interactions into executable robot data for policy learning.

\subsection{Human-to-robot Transfer and Physical Realization}
Human-to-robot transfer methods convert human hand--object demonstrations into robot motion references while accounting for differences in hand morphology~\cite{22eccv_dexmv,22arxiv_dextransfer,25cvpr_maniptrans,25arxiv_hermes, 25arxiv_spider,26arxiv_deximit,26arxiv_doasido}.
Interaction-mesh-based methods preserve local spatial relationships between the body and surrounding objects during motion retargeting~\cite{12humanoids_interactionmesh,25arxiv_omniretarget}.
Concurrent work applies related formulations to dexterous hands to preserve hand--object geometry and contact relationships~\cite{26arxiv_toporetarget,26arxiv_c2dex}.
Physics-based imitation and residual control further turn these motion references into executable robot behaviors~\cite{25cvpr_maniptrans,26arxiv_regrind,26arxiv_c2dex}.
Building on these techniques, we adapt each human source sequence to compatible object shapes and different arm trajectories to generate robot demonstrations that retain the demonstrated grasp.

\subsection{Robot-centric Data Generation for Policy Learning}
Another line of work studies how to scale robot training data beyond direct teleoperation \cite{23corl_mimicgen} for policy training \cite{23ijrr_diffusionpolicy, 23rss_act, 24arxiv_dp3} and VLA training \cite{24arxiv_pi0, 26arxiv_lingbotvla}.
Simulation-based augmentation methods vary object configurations, task layouts, and motion trajectories to produce data for visuomotor policy learning~\cite{25icra_dexmimicgen,25rss_dex1b}.
UniDex uses human-in-the-loop retargeting to convert human demonstrations into robot trajectories and trains a VLA on explicit 3D point-cloud observations~\cite{26arxiv_unidex}.
Starting from selected human sequences, our pipeline automatically retargets and augments the interactions, realizes them through physics-based imitation, and renders RGB observations for VLA learning.
We further organize the source interactions by grasp mode and pair the generated demonstrations with descriptions of how to grasp.

\section{Method}
\label{sec:method}
\begin{figure*}[!t]
    \begin{center}
        \includegraphics[width=0.9\linewidth]{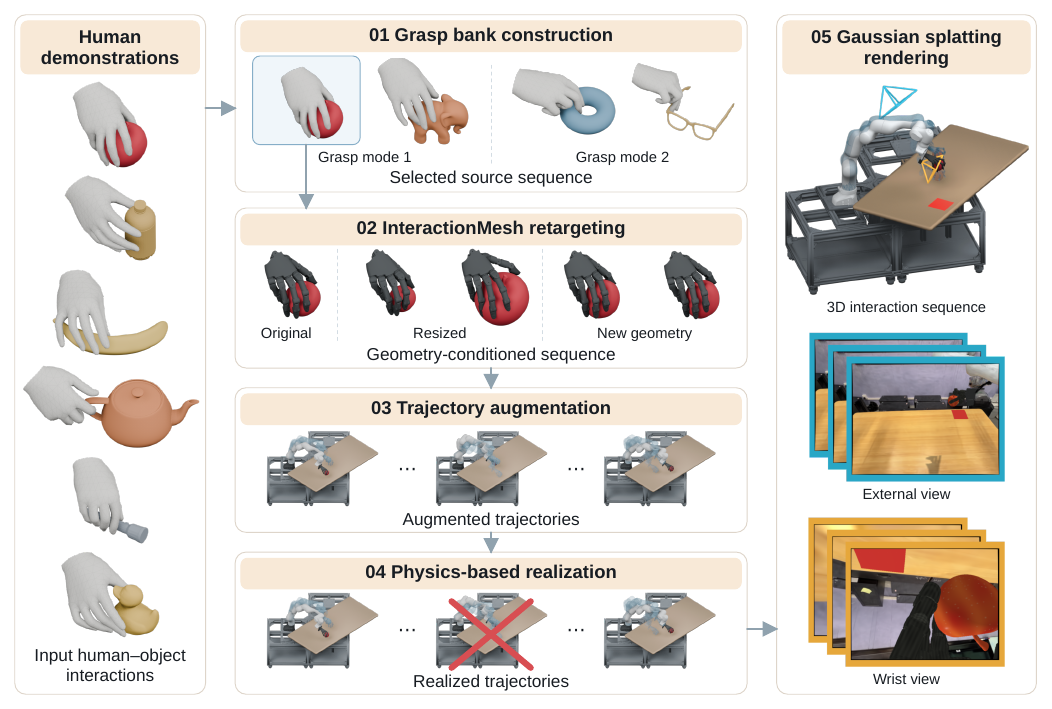}
    \end{center}
    \caption{\textbf{\shortname{} Overview}.
    (01) We organize human demonstrations by grasp mode and select interaction seeds.
(02) InteractionMesh retargeting adapts these seeds to the robot hand and new object geometries.
(03) Trajectory augmentation adds whole-arm motion variations, and (04) physics-based realization produces executable rollouts.
(05) Gaussian Splatting rendering provides visual observations, which are paired with robot actions and grasp descriptions for VLA training.}
    \label{fig:pipeline}
\end{figure*}

Given a collection of human hand--object interaction sequences, our goal is to generate diverse, physically executable robot demonstrations for dexterous policy learning.
As illustrated in \cref{fig:pipeline}, we first organize human interactions by grasp mode and assign corresponding language descriptions (\cref{sec:grasp_bank}).
We then transform selected human sequences into diverse, physically executable robot demonstrations (\cref{sec:dex_data}).
Finally, we use the generated data to fine-tune a VLA for coordinated arm and finger control following language instructions (\cref{sec:vla_train}).

\subsection{Grasp Bank Construction}
\label{sec:grasp_bank}
Human hand--object datasets~\cite{20eccv_grab} contain recurring grasp poses across different object instances and motion sequences.
We construct a grasp bank to organize these patterns, select representative source interactions, and provide consistent grasp-mode annotations for the generated robot data.

For each sequence, we select the first frame at which the object's displacement from its initial position exceeds a small threshold.
We use this frame as an estimate of grasp establishment and compute a 170-dimensional descriptor:
\begin{equation}
\mathbf{f}
= \operatorname{concat}(
\mathbf{a}, \mathbf{x}, \mathbf{g}, \mathbf{h},
d, \mathbf{s}, \mathbf{e}, c
)
\in \mathbb{R}^{170},
\label{eq：grasp_descriptor}
\end{equation}
where $\operatorname{concat}$ concatenates eight feature blocks.
$\mathbf{a}\in\mathbb{R}^{13}$ describes how much the fingers bend, which contains the sum of joint-bending angles per finger, five fingertip-to-palm distances, and three measures of hand opening.
The last three values are the mean, standard deviation, and maximum of the distances from the five fingertips to their centroid.
$\mathbf{x}\in\mathbb{R}^{63}$ contains the positions of 21 MANO \cite{17sigasia_mano} joints relative to the palm.
$\mathbf{g}\in\mathbb{R}^{3}$ records the world-up direction in the palm's local frame.
$\mathbf{h}\in\mathbb{R}^{21}$ contains the vertical offsets of the hand landmarks from the centroid of sampled object surface points.
$d\in\mathbb{R}$ measures the palm's vertical offset from its nearest object surface point.
$\mathbf{s}\in\mathbb{R}^{63}$ contains vectors from the 21 hand joints to their nearest object surface points, expressed in the same frame as $\mathbf{x}$.
$\mathbf{e}\in\mathbb{R}^{5}$ measures whether each finger has object surfaces on both sides, as when passing through a handle.
For $c\in\mathbb{R}$, we fit a plane to the contact points and measure how directly it faces the object center.
Specifically, $c$ is the squared cosine of the angle between the plane's normal and the line joining the contact center and object center.
We compute distances between grasp descriptors across all sequences and group similar grasps into modes using a fixed threshold.

From these groups, we manually select eight representative grasp modes and choose source sequences within each mode as interaction seeds.
For each mode, GPT~\cite{22arxiv_chatgpt} generates a description of the grasp.
All robot demonstrations generated from that mode share the same description.

\subsection{Dexterous Robot Data Generation}
\label{sec:dex_data}

\subsubsection{InteractionMesh-based Cross-Embodiment and Cross-Object Retargeting}
Human and robot hands differ in morphology, and point-wise landmark matching alone does not explicitly model their interaction with the object.
To transfer a demonstrated interaction across embodiments and compatible object geometries, we use InteractionMesh~\cite{12humanoids_interactionmesh,25arxiv_omniretarget} to encode local hand--object spatial relationships.

Given a human hand--object sequence, we express the interaction in the source object's local coordinate frame.
We represent the human hand using $21$ MANO \cite{17sigasia_mano} landmarks covering the wrist, finger joints, and fingertips, and sample anchors on the source object surface.
At a representative frame within the interaction interval, we jointly tetrahedralize the hand landmarks and object anchors using 3D Delaunay triangulation.
The resulting connectivity defines a fixed neighborhood $\mathcal{N}(i)$ for each vertex throughout the sequence.
For vertex $\mathbf{v}_{t,i}$ at frame $t$, its Laplacian coordinate is
\begin{equation}
    \boldsymbol{\delta}_{t,i}
    =
    \mathbf{v}_{t,i}
    -
    \frac{1}{|\mathcal{N}(i)|}
    \sum_{j\in\mathcal{N}(i)}
    \mathbf{v}_{t,j}.
    \label{eq:interactionmesh}
\end{equation}
These coordinates describe each vertex relative to its neighboring hand and object vertices, providing a local representation of the demonstrated interaction.

For cross-object transfer, we extract multi-view surface features using Diff3F~\cite{24cvpr_diff3f} and match features to establish point correspondences between the source and target objects.
We estimate a similarity transformation, including rotation, translation, and scale, from these correspondences to align the target object with the source and bring it to a comparable size.
The aligned correspondences associate each source InteractionMesh anchor with a point on the target surface, obtaining paired anchor matrices $\mathbf{P}_{\mathrm{src}}$ and $\mathbf{P}_{\mathrm{tgt}}$.
The aligned target geometry follows the source object's trajectory, providing a kinematic reference for the transferred interaction.

We use the Wuji~\cite{wuji} dexterous robot hand as the target embodiment.
At each frame, the robot configuration $q_t$ determines the corresponding hand landmark positions through forward kinematics.
Using \cref{eq:interactionmesh}, we compute $\boldsymbol{\delta}^{\mathrm{src}}_{t,i}$ for the human--source interaction and $\boldsymbol{\delta}^{\mathrm{tgt}}_{t,i}$ for the robot--target interaction.
Both use the same fixed neighborhood connectivity, with all vertex positions expressed in the shared object-local frame.
We optimize the robot hand trajectory over $T$ frames by matching the Laplacian coordinates of hand and object vertices:
\begin{equation}
    \begin{aligned}
    \mathcal{L}_{\mathrm{InterMesh}}
    = \sum_{t=1}^{T} \Biggl(
    &\sum_{i \in \mathcal{V}_{H}}
    \left\|
        \boldsymbol{\delta}^{\mathrm{tgt}}_{t,i}
        - \boldsymbol{\delta}^{\mathrm{src}}_{t,i}
    \right\|_2^2 \\
    +{}&\sum_{i \in \mathcal{V}_{O}}
    \left\|
        \boldsymbol{\delta}^{\mathrm{tgt}}_{t,i}
        - \boldsymbol{\delta}^{\mathrm{src}}_{t,i}
    \right\|_2^2
    \Biggr),
    \end{aligned}
    \label{eq:interactionmesh_loss}
\end{equation}
where $\mathcal{V}_{H}$ and $\mathcal{V}_{O}$ denote the hand and object vertex index sets, respectively.
The target object anchors remain fixed, while the robot wrist pose and finger joints are optimized to preserve the local interaction structure.
Object-vertex Laplacian coordinates also depend on neighboring hand landmarks and therefore contribute to the optimization.
When the source object is also used as the target, the same formulation performs human-to-robot retargeting on the original object.

We combine $\mathcal{L}_{\mathrm{InterMesh}}$ with contact-aware fingertip constraints, soft penalties for hand--object and self-penetration, morphology-aware hand-pose constraints, and temporal regularization for smooth wrist and finger motion.
We optimize the combined objective over the robot wrist pose and 20-DoF finger configuration throughout the sequence, independently for each target geometry.
The resulting hand--object trajectories serve as candidate references for subsequent whole-arm trajectory augmentation and physics-based imitation.

\subsubsection{Whole-Arm Trajectory Augmentation}
The retargeting stage provides robot hand--object motions and then we augment these motions to create multiple tabletop manipulation sequences.
Since the GRAB sequences used in our pipeline primarily involve pick-and-place interactions~\cite{20eccv_grab}, we vary where the hand starts, where the object is initially placed, and where it is moved.

For each variant, we sample an initial wrist pose within the robot's workspace, an object pose on the table, and a placement target.
We retain the retargeted grasp-and-lift motion and translate and rotate the hand and object together to match the sampled object pose.
An approach trajectory connects the initial wrist pose to the beginning of this motion, matching position and velocity at the connection.
We then move the wrist and object together toward the placement target while retaining the reference finger motion.
At the placement target, we reuse the source motion to set down and release the object, then withdraw the hand.
This produces a hand--object trajectory covering approach, grasping, transport, placement, and retreat.
Finally, we solve inverse kinematics~\cite{mink} along the wrist trajectory to obtain arm joint references and discard variants for which no feasible solution is found.

\subsubsection{Physics-Based Dexterous Imitation}
The augmented pick-and-place trajectories provide diverse kinematic references, but direct replay does not guarantee successful execution under contact dynamics, where tracking errors can lead to missed contacts, object slip, or grasp failure.
We therefore use closed-loop physics-based imitation~\cite{18sig_deepmimic} to obtain executable robot rollouts.
For each source sequence, we train a separate policy jointly on its augmented references, covering variations in object geometry and whole-arm pick-and-place trajectories.

We separate whole-arm motion from dexterous finger control: the robot arm tracks the precomputed joint-space reference trajectories using PD control, while the learned policy controls only the 20-DoF robot fingers.
The policy observation includes current and reference finger states, actual and reference wrist-relative object poses, a BPS encoding of the target-object geometry~\cite{19iccv_bps,25corl_dexplore}, reference hand--object interaction and contact cues, measured fingertip contact forces, short-horizon future references, and whole-arm dynamic context including wrist-tracking errors.
These inputs are concatenated into a 542-dimensional observation vector.
Given this vector, the policy predicts a 20-dimensional residual action, which is added to the reference finger-joint targets \cite{25cvpr_maniptrans}.
The arm joint targets are taken from the precomputed references and are not modified by the policy.

We train the policy with a reward that balances object tracking, interaction preservation, and smooth finger control:
\begin{equation}
\begin{aligned}
R ={}&
0.40 R_{\mathrm{obj}}
+ 0.30 R_{\mathrm{contact}}
+ 0.20 R_{\mathrm{tip}}
+ 0.10 R_{\mathrm{joint}} \\
&- 0.025 P_{\mathrm{smooth}}
- 0.01 P_{\mathrm{act}}
- 0.05 P_{\mathrm{cross}}.
\end{aligned}
\label{eq:reward}
\end{equation}
Here, $R_{\mathrm{contact}}$ encourages contact by the fingers involved in the reference interaction, $R_{\mathrm{tip}}$ preserves the corresponding fingertip locations in the object frame, and $R_{\mathrm{joint}}$ tracks the reference finger articulation.
The penalties $P_{\mathrm{smooth}}$, $P_{\mathrm{act}}$, and $P_{\mathrm{cross}}$ discourage abrupt target changes, unnecessarily large residual actions, and fingertip--fingertip collisions, respectively.
Since the finger policy does not control the arm, object-tracking errors may partly arise from imperfect tracking of the precomputed arm commands.
We therefore introduce a wrist-error-dependent tolerance in the object-tracking reward $R_{\mathrm{obj}}$.
The reward reaches its maximum when the object-pose error lies within this tolerance and decreases beyond it, avoiding strict penalties for deviations within the allowance determined by arm-tracking error.

After training, we evaluate each source-specific policy through simulation rollouts and retain only executions that satisfy the task and physical-success criteria.
Successful rollouts are recorded as robot demonstrations for subsequent rendering and downstream VLA learning.

\subsubsection{Robot-Centric Rendering}
To render realistic visual observations, we capture a real-world workspace and reconstruct it using 3D Gaussian Splatting~\cite{23tog_gs}.
Within this reconstructed scene, we replay successful simulation rollouts and render synchronized images from head-view and wrist-mounted cameras.
We augment lighting conditions, object colors and textures, and shadow appearance, and apply small perturbations to the intrinsic and extrinsic parameters of both cameras.
These augmentations introduce visual diversity around the deployment setting while keeping the underlying robot and object trajectories and action labels unchanged.
Each rendered timestep is paired with the corresponding robot state and action to form visual--action demonstrations for VLA learning.

\subsection{Robot-Centric VLA Learning}
\label{sec:vla_train}
We aggregate demonstrations across all source sequences, with trajectories from the same grasp mode sharing a language instruction describing the intended finger configuration (\cref{sec:grasp_bank}).
The resulting multimodal dataset is used to fine-tune a pretrained Lingbot-VLA model~\cite{26arxiv_lingbotvla} for language-conditioned pick-and-place manipulation.
At deployment, the VLA takes head-view and wrist-camera images, robot proprioception, and a language instruction specifying the desired grasping strategy, and jointly predicts actions for the seven arm joints and 20 finger joints.

\section{Experiments}
\label{sec:experiments}

\begin{table}[t]
    \centering
    \caption{
        Dataset composition by reference grasp mode.
        G1--G8 identify the grasp modes throughout the paper.
    }
    \label{tab:dataset_composition}
    \small
    \setlength{\tabcolsep}{3pt}
    \begin{tabular}{@{}llcc@{}}
        \toprule
        ID & Reference grasp mode
        & \shortstack{Source\\sequences}
        & \shortstack{Robot\\trajectories} \\
        \midrule
        G1 & Oblique side whole-hand grasp   & 5 & 6,596 \\
        G2 & Top-down deep grip        & 2 & 2,559 \\
        G3 & Top-down fingertip grasp  & 2 & 2,475 \\
        G4 & Top-down whole-hand grasp & 2 & 2,481 \\
        G5 & Handle-threading grasp    & 2 & 3,452 \\
        G6 & High-wrist oblique grasp  & 2 & 2,680 \\
        G7 & Lateral body wrap         & 2 & 2,442 \\
        G8 & Two-finger hook    & 2 & 2,585 \\
        \midrule
        \multicolumn{2}{@{}l}{Total} & 19 & 25,270 \\
        \bottomrule
    \end{tabular}
\end{table}

\subsection{Evaluation Setup}
\label{sec:exp_setup}
We select \graspmodenum{} representative grasp modes from GRAB~\cite{20eccv_grab} and use \grabseqnum{} distinct human hand--object sequences spanning 17 object categories.
The resulting dataset contains \vlaseqnum{} physically verified robot trajectories, with the per-mode composition summarized in \cref{tab:dataset_composition}.

For physics-based imitation, we use an MLP with four hidden layers of $1,024$ units each to predict $20$-dimensional residual finger actions around the reference joint targets.
The policy is trained with a learning rate of $2 \times 10^{-5}$ using $4,096$ parallel simulation environments for $1500$ iterations in isaac-lab \cite{isaaclab}.
The policy control and simulation frequencies are $30$\,Hz and $120$\,Hz, respectively.
For VLA training, we adopt the LingBot-VLA architecture~\cite{26arxiv_lingbotvla} and adapt its action output layer and state input projection to support 27-dimensional actions and robot states, corresponding to the seven joints of the Franka arm \cite{franka} and $20$ finger joints.
In our single-arm setup, the policy takes images from a head-view camera and a wrist-mounted camera, together with the robot state and a grasp-mode language instruction, as input.
We fine-tune the model for $55,000$ steps on eight H20 GPUs, using a batch size of $256$ and a learning rate of $5 \times 10^{-5}$.
At deployment, the VLA runs inference at approximately $5$\,Hz and predicts 50-step action chunks.
We use RTC~\cite{26nips_rtc} to execute 30 steps from each chunk before switching to the next.

We evaluate retargeting quality and physics-based imitation against their respective baselines.
Unless otherwise specified, these comparisons use the same robot embodiment, object assets, task configurations, and evaluation protocols.
We further evaluate the trained VLA's language-conditioned manipulation performance in simulation and on a real robot.

\subsection{Interaction Retargeting}
\label{sec:exp:retargeting}
We compare our retargeting method with Dex-Retargeting~\cite{23rss_anyteleop}, SPIDER~\cite{25arxiv_spider}, and the official Wuji retargeter~\cite{wuji}, using the same \grabseqnum{} human source sequences and robot embodiment.
We report fingertip contact error, contact angular error, and average penetration depth.

\begin{table}[t]
    \centering
    \caption{Human-to-robot interaction retargeting.}
    \label{tab:retargeting_results}
    \small
    \setlength{\tabcolsep}{3pt}
    \begin{tabular}{@{}lccc@{}}
        \toprule
        Method
        & \shortstack{Contact error\\(mm) $\downarrow$}
        & \shortstack{Angular error\\($^\circ$) $\downarrow$}
        & \shortstack{Penetration\\(mm) $\downarrow$} \\
        \midrule
        Dex-Retargeting
        & 9.486 & 18.560 & 3.870 \\
        SPIDER
        & 3.521 & 16.665 & 3.418 \\
        Wuji official
        & 5.273 & 16.350 & \textbf{2.959} \\
        Ours
        & \textbf{3.326} & \textbf{7.636} & 3.371 \\
        \bottomrule
    \end{tabular}
\end{table}

As shown in \cref{tab:retargeting_results}, our method outperforms all three baselines in contact accuracy and angular alignment, indicating better preservation of the demonstrated hand--object interactions.
\begin{figure}[t]
    \centering
    \includegraphics[width=1.0\columnwidth]{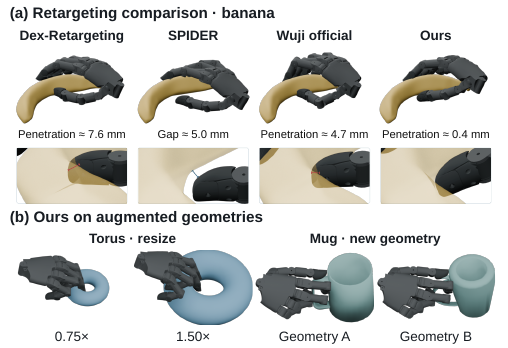}
    \caption{
    Qualitative retargeting results.
    (a) Comparison with baseline retargeting methods.
    Gap marks a separation between the hand and object surface.
    (b) Cross-object transfer to torus objects scaled by $0.75\times$ and $1.5\times$, and to mugs with different geometries.
    }
    \label{fig:retarget_vis}
\end{figure}

The qualitative comparison in \cref{fig:retarget_vis}(a) further illustrates the improved preservation of the reference grasp configurations.
In addition, \cref{fig:retarget_vis}(b) shows cross-object retargeting to scaled torus objects ($0.75\times$ and $1.5\times$) and mugs with different geometries, where the robot hand adapts to the target geometry while retaining the intended grasping.

\subsection{Physics-Based Dexterous Imitation}
\label{sec:exp:mimic}
We next evaluate how reliably the generated kinematic references can be realized under contact dynamics.
We compare our physics-based imitation with ManipTrans~\cite{25cvpr_maniptrans} and Dexplore~\cite{25corl_dexplore} on the same manipulation sequences.
The baselines are trained with a floating hand, whereas our finger policy is trained with arm tracking in the loop.
Our primary evaluation uses the FR3 arm \cite{franka} to track the reference wrist trajectory.
We report the error in the reference wrist--object distance relation, object rotation error, contact recall, and rollout success rate.
Contact recall measures the fraction of reference-contact frames in which the corresponding physical contact is established.
Tracking errors and contact recall include both successful and failed rollouts.
For comparison, we also report floating-hand results for the baselines and our w/o Arm variant, which is trained with a floating hand.

\begin{table}[t]
    \centering
    \caption{
        Physics-based realization of kinematic references.
        FR3 denotes evaluation with arm tracking, and Float
        denotes floating-hand evaluation.
        Contact denotes contact recall.
        The w/o Arm variant is trained with a virtual wrist.
    }
    \label{tab:mimic_results}
    \setlength{\tabcolsep}{3pt}
    \begin{tabular}{lccccc}
        \toprule
        & \multicolumn{4}{c}{FR3} & Float \\
        \cmidrule(lr){2-5}
        \cmidrule(lr){6-6}
        Method
        & \shortstack{Obj. Dist.\\(cm) $\downarrow$}
        & \shortstack{Obj. Rot.\\($^\circ$) $\downarrow$}
        & \shortstack{Contact\\(\%) $\uparrow$}
        & \shortstack{Success\\(\%) $\uparrow$}
        & \shortstack{Success\\(\%) $\uparrow$} \\
        \midrule
        ManipTrans
        & 12.42 & 25.25 & 32.24 & 34.25 & 71.32 \\
        Dexplore
        & 14.84 & 33.74 & 33.87 & 24.81 & 94.33 \\
        Ours
        & \textbf{2.61}
        & \textbf{7.71}
        & \textbf{78.75}
        & \textbf{95.50}
        & -- \\
        \midrule
        w/o Arm
        & 13.65 & 28.83 & 30.50 & 28.02 & 98.04 \\
        \bottomrule
    \end{tabular}
\end{table}

As shown in \cref{tab:mimic_results}, our method achieves lower tracking errors, higher contact recall, and higher success under arm tracking than both baselines.
The baselines and w/o Arm ablation perform well with a floating hand but become less reliable when coupled with the robot arm.
This gap highlights the importance of training the finger controller with arm-tracking dynamics for reliable whole-arm execution.

\subsection{VLA Learning in Simulation}
\label{sec:exp:vla_sim}
We evaluate whether the generated robot demonstrations provide effective supervision for downstream VLA learning.
We focus on pick-and-place manipulation, where a rollout is considered successful if the robot lifts the target object from the table and places it inside the designated goal region.
Unless otherwise stated, all comparisons use the same VLA architecture and training protocol to assess the effect of the training data.
Our evaluation covers all eight grasp modes using their corresponding language instructions and representative objects.
We construct four evaluation settings that progressively introduce unseen factors.
The first uses object geometries, layouts, and appearances represented in the training data.
The second introduces unseen layouts while retaining seen geometries and appearances.
The third additionally introduces unseen object geometries, and the fourth further introduces unseen object appearances.
Here, layout specifies the object's initial position and target placement region, geometry refers to the object shape, and appearance refers to visual properties such as object color.
For each setting, we evaluate ten trials per grasp mode with different start and placement configurations, yielding $80$ trials per setting and $320$ trials in total.
All comparisons use the same evaluation suite, with seen and unseen conditions defined relative to the full training set.

We compare VLA training on our data with training on kinematic references and the data generated using ManipTrans \cite{25cvpr_maniptrans} and Dexplore \cite{25corl_dexplore}.
The kinematic references are rendered directly without physics-based realization, while the ManipTrans and Dexplore rollouts are processed through the same rendering and VLA training stages as ours.
We separately remove geometry augmentation, rendering augmentation, and grasp descriptions.
The text ablation retains task instructions but omits descriptions of how to grasp.
We also fine-tune $\pi_0$~\cite{24arxiv_pi0} on our full dataset to test its utility with another VLA architecture.

\begin{table}[t]
    \centering
    \caption{
    VLA pick-and-place success rates (\%) in simulation
    using different training datasets and policy variants.
    L, G, and A denote layout, object geometry, and appearance.
    The w/o Grasp Mode Text variant omits grasp-specific text.
    The w/ $\pi_0$ variant fine-tunes $\pi_0$ on the same dataset.
    }
    \label{tab:vla_sim_results}
    \setlength{\tabcolsep}{4pt}
    \begin{tabular}{lcccc}
        \toprule
        & & \multicolumn{3}{c}{Unseen Factors} \\
        \cmidrule(lr){3-5}
        Method & Seen & L & L+G & L+G+A \\
        \midrule
        Kinematic Data   & 7.50  & 12.50 & 7.50  & 5.00  \\
        ManipTrans Data  & 38.75 & 43.75 & 38.75 & 51.25 \\
        Dexplore Data    & 32.50 & 40.00 & 26.25 & 32.50 \\
        Ours
        & \textbf{97.50} & \textbf{95.00}
        & \textbf{91.25} & \textbf{91.25} \\
        \midrule
        w/o Geo. Aug.    & 92.50 & 93.75 & 75.00 & 76.25 \\
        w/o Render. Aug. & 91.25 & 88.75 & 87.50 & 76.25 \\
        w/o Grasp Mode Text   & 80.00 & 80.00 & 76.25 & 70.00 \\
        w/ $\pi_0$       & 88.75 & 88.75 & 86.25 & 90.00 \\
        \bottomrule
    \end{tabular}
\end{table}

As shown in \cref{tab:vla_sim_results}, our demonstrations yield higher task success than kinematic data and data generated using either imitation baseline across all evaluation settings.
The poor performance of kinematic data highlights the importance of realizing references under contact dynamics before using them for policy learning.
Removing geometry augmentation causes a larger performance drop when unseen geometries are introduced, while removing rendering augmentation has the greatest effect when appearance also changes.
Omitting grasp descriptions reduces task success across all settings.
This suggests that pairing grasp modes with distinct language descriptions helps the policy reproduce the intended finger configurations.
Fine-tuning $\pi_0$ \cite{24arxiv_pi0} on the same data yields comparable results, although LingBot-VLA~\cite{26arxiv_lingbotvla} achieves higher success rates.

\begin{table}[t]
    \centering
    \caption{
    Grasp instruction control on the same object.
    G1 and G4 denote the grasp modes.
    Bold values indicate the closer reference in each row.
    }
    \label{tab:grasp_instruction}
    \setlength{\tabcolsep}{4pt}
    \begin{tabular}{llcc}
        \toprule
        & & \multicolumn{2}{c}{MAE to reference ($^\circ$)} \\
        \cmidrule(lr){3-4}
        Policy & Instruction & G1 & G4 \\
        \midrule
        \multirow{2}{*}{Ours}
        & G1 & \textbf{4.45} & 15.31 \\
        & G4 & 14.46 & \textbf{4.33} \\
        \midrule
        w/o Grasp Mode Text & -- & \textbf{4.37} & 15.03 \\
        \bottomrule
    \end{tabular}
\end{table}

To examine whether language instructions control the grasp, we evaluate two instructions, G1 and G4, on the same apple object.
For trials with valid lifting intervals, we compute MAE across the $20$ finger joints using median angles during sustained lifting.
Each executed grasp is compared with both references.
As shown in \cref{tab:grasp_instruction}, all ten trials per instruction are closer to the requested reference, whereas all ten task-only trials without grasp descriptions favor G1.
These results support language-based selection between grasps of the same object.

\begin{table}[t]
    \centering
    \caption{
        Effect of training-data scale.
        All data ratios use the same evaluation settings.
    }
    \label{tab:vla_data_scale}
    \setlength{\tabcolsep}{4pt}
    \begin{tabular}{lcccc}
        \toprule
        & & \multicolumn{3}{c}{Unseen Factors} \\
        \cmidrule(lr){3-5}
        Data Ratio & Seen & L & L+G & L+G+A \\
        \midrule
        1\%   & 67.50 & 38.75 & 31.25 & 28.75 \\
        10\%  & 85.00 & 87.50 & 90.00 & 85.00 \\
        100\% & \textbf{97.50} & \textbf{95.00} & \textbf{91.25} & \textbf{91.25} \\
        \bottomrule
    \end{tabular}
\end{table}

We next study how VLA performance scales with the amount of generated data.
We uniformly subsample different fractions of the full training set and independently train the VLA on each subset using the same training protocol and evaluation suite.
As shown in \cref{tab:vla_data_scale}, training on the smallest subset substantially reduces task success, especially under unseen conditions.

\subsection{Real-World VLA Deployment}
\label{sec:exp:vla_real}
\begin{figure}[t]
    \centering
    \includegraphics[width=1.0\columnwidth]{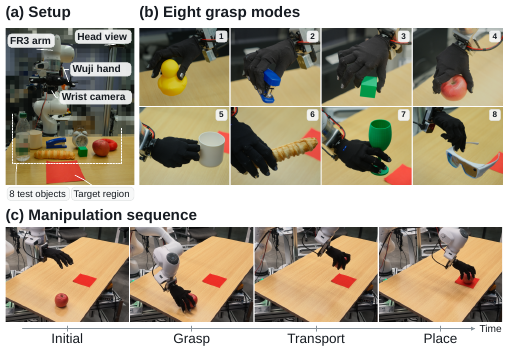}
    \caption{
    \textbf{Real-world VLA deployment.}
(a) Robot setup and test objects.
(b) Grasping various objects using eight grasp modes.
(c) A representative manipulation sequence executed by the learned VLA.}
    \label{fig:quali}
\end{figure}

Finally, we evaluate the VLA trained on our data on the real robot.
We use a Franka arm equipped with a Wuji dexterous hand and evaluate all eight grasp modes, using one test object per mode and conducting ten trials for each object--mode pair.
The policy receives head-view and wrist-camera observations together with the associated grasp instruction and jointly predicts arm and finger actions.
\cref{fig:quali} (a) shows the deployment setup and test objects.
A trial is considered successful if the robot grasps and lifts the target object, then places it inside the goal region.

\cref{fig:quali} (b) presents grasp executions across different objects using the eight grasp modes.
\cref{fig:quali} (c) further visualizes successive stages of a representative VLA execution, illustrating how the policy coordinates arm motion and finger control.

\begin{table}[t]
    \centering
    \caption{
        Real-world VLA success rates.
    }
    \label{tab:vla_real_results}
    \setlength{\tabcolsep}{3pt}
    \begin{tabular}{@{}l*{8}{c}@{}}
        \toprule
        Grasp Mode
        & G1 & G2 & G3 & G4 & G5 & G6 & G7 & G8 \\
        \midrule
        Success (\%) $\uparrow$
        & 60 & 40 & 70 & 70 & 20 & 50 & 50 & 80 \\
        \bottomrule
    \end{tabular}
\end{table}

As shown in \cref{tab:vla_real_results}, the VLA trained on our generated demonstrations successfully executes all eight grasp modes on the real robot.
Success rates vary across modes, with handle threading achieving the lowest success, consistent with its tighter requirements for precise finger--object alignment.

\subsection{Limitations and Future Work}
Our current study focuses on representative grasp modes and primarily evaluates grasping and pick-and-place tasks.
Future work could extend this approach to in-hand manipulation, where the fingers change contact as the object moves within the hand.
Extending the framework to bimanual manipulation would further support coordinated tasks, such as object handovers and operations where one hand stabilizes an object while the other manipulates it.
Generating demonstrations for robot hands with different finger structures and joint configurations could also help study how learned grasping behaviors transfer across embodiments.
Finally, adding tactile observations to the generated data and VLA inputs could help policies detect slip and adjust finger contact during execution.

\section{Conclusion}
\label{sec:conclusion}
We present \shortname{}, a framework that transforms human hand--object interactions into diverse, physically executable robot data for language-guided dexterous manipulation.
The framework organizes human interactions by grasp mode and expands individual sequences across object geometries and arm trajectories through interaction-preserving retargeting and physics-based imitation.
The resulting demonstrations are paired with robot-centric observations and shared descriptions of how to grasp.
Experiments show improved contact fidelity and more reliable physical realization than the evaluated baselines.
VLA evaluation in simulation and deployment on a real robot further demonstrate that the generated data support learning coordinated arm and finger actions from visual observations and language instructions.

{\small
\bibliographystyle{IEEEtran}
\bibliography{main}

@article{17sigasia_mano,
  title         = {Embodied Hands: Modeling and Capturing Hands and Bodies Together},
  author        = {Romero, Javier and Tzionas, Dimitrios and Black, Michael J.},
  journal       = {ACM Trans. Graph.},
  volume        = {36},
  number        = {6},
  series        = {245:1--245:17},
  month         = nov,
  year          = {2017},
  month_numeric = {11}
}

@article{18sig_deepmimic,
  title={Deepmimic: Example-guided deep reinforcement learning of physics-based character skills},
  author={Peng, Xue Bin and Abbeel, Pieter and Levine, Sergey and Van de Panne, Michiel},
  journal={ACM Transactions On Graphics (TOG)},
  volume={37},
  number={4},
  pages={1--14},
  year={2018},
  publisher={ACM New York, NY, USA}
}

@inproceedings{19iccv_bps,
  title={Efficient learning on point clouds with basis point sets},
  author={Prokudin, Sergey and Lassner, Christoph and Romero, Javier},
  booktitle={Proceedings of the IEEE/CVF international conference on computer vision},
  pages={4332--4341},
  year={2019}
}

@inproceedings{20eccv_grab,
  title={GRAB: A dataset of whole-body human grasping of objects},
  author={Taheri, Omid and Ghorbani, Nima and Black, Michael J and Tzionas, Dimitrios},
  booktitle={Computer Vision--ECCV 2020: 16th European Conference, Glasgow, UK, August 23--28, 2020, Proceedings, Part IV 16},
  pages={581--600},
  year={2020},
  organization={Springer}
}

@inproceedings{22cvpr_oakink,
  author    = {Yang, Lixin and Li, Kailin and Zhan, Xinyu and Wu, Fei and Xu, Anran and Liu, Liu and Lu, Cewu},
  title     = {{OakInk}: A Large-Scale Knowledge Repository for Understanding Hand-Object Interaction},
  booktitle = {IEEE/CVF Conference on Computer Vision and Pattern Recognition (CVPR)},
  year      = {2022},
}

@misc{22arxiv_chatgpt,
  title        = { OpenAI: Introducing ChatGPT},
  author       = {OpenAI},
  year         = {2022},
  howpublished = {\url{https://openai.com/blog/chatgpt}}
}

@inproceedings{23corl_mimicgen,
    title={MimicGen: A Data Generation System for Scalable Robot Learning using Human Demonstrations},
    author={Mandlekar, Ajay and Nasiriany, Soroush and Wen, Bowen and Akinola, Iretiayo and Narang, Yashraj and Fan, Linxi and Zhu, Yuke and Fox, Dieter},
    booktitle={7th Annual Conference on Robot Learning},
    year={2023}
}

@InProceedings{24cvpr_oakink2,
    author    = {Zhan, Xinyu and Yang, Lixin and Zhao, Yifei and Mao, Kangrui and Xu, Hanlin and Lin, Zenan and Li, Kailin and Lu, Cewu},
    title     = {OAKINK2: A Dataset of Bimanual Hands-Object Manipulation in Complex Task Completion},
    booktitle = {Proceedings of the IEEE/CVF Conference on Computer Vision and Pattern Recognition (CVPR)},
    month     = {June},
    year      = {2024},
    pages     = {445-456}
}

@inproceedings{20icra_dexpilot,
  title={Dexpilot: Vision-based teleoperation of dexterous robotic hand-arm system},
  author={Handa, Ankur and Van Wyk, Karl and Yang, Wei and Liang, Jacky and Chao, Yu-Wei and Wan, Qiang and Birchfield, Stan and Ratliff, Nathan and Fox, Dieter},
  booktitle={IEEE International Conference on Robotics and Automation},
  year={2020}
}

@article{22arxiv_dextransfer,
  title={Dextransfer: Real world multi-fingered dexterous grasping with minimal human demonstrations},
  author={Chen, Zhongyu Q. and Van Wyk, Karl and Chao, Yu-Wei and Yang, Wei and Mousavian, Arsalan and Gupta, Abhishek and Fox, Dieter},
  journal={arXiv preprint arXiv:2209.14284},
  year={2022}
}

@inproceedings{22eccv_dexmv,
  title={Dexmv: Imitation learning for dexterous manipulation from human videos},
  author={Qin, Yuzhe and Wu, Yueh-Hua and Liu, Shaowei and Jiang, Hanwen and Yang, Ruihan and Fu, Yang and Wang, Xiaolong},
  booktitle={European Conference on Computer Vision},
  pages={570--587},
  year={2022}
}

@inproceedings{23rss_anyteleop,
  title={{AnyTeleop}: A general vision-based dexterous robot arm-hand teleoperation system},
  author={Qin, Yuzhe and Yang, Wei and Huang, Binghao and Van Wyk, Karl and Su, Hao and Wang, Xiaolong and Chao, Yu-Wei and Fox, Dieter},
  booktitle={Robotics: Science and Systems},
  year={2023}
}

@inproceedings{23rss_act,
  title={Learning fine-grained bimanual manipulation with low-cost hardware},
  author={Zhao, Tony Z. and Kumar, Vikash and Levine, Sergey and Finn, Chelsea},
  booktitle={Robotics: Science and Systems},
  year={2023}
}

@article{23ijrr_diffusionpolicy,
  title={Diffusion policy: Visuomotor policy learning via action diffusion},
  author={Chi, Cheng and Xu, Zhenjia and Feng, Siyuan and Cousineau, Eric and Du, Yilun and Burchfiel, Benjamin and Tedrake, Russ and Song, Shuran},
  journal={The International Journal of Robotics Research},
  pages={02783649241273668},
  year={2023}
}

@article{24arxiv_dp3,
  title={3D diffusion policy: Generalizable visuomotor policy learning via simple 3D representations},
  author={Ze, Yanjie and Zhang, Gu and Zhang, Kangning and Hu, Chenyuan and Wang, Muhan and Xu, Huazhe},
  journal={arXiv preprint arXiv:2403.03954},
  year={2024}
}

@article{24arxiv_dexcap,
  title={Dexcap: Scalable and portable mocap data collection system for dexterous manipulation},
  author={Wang, Chen and Shi, Haochen and Wang, Weizhuo and Zhang, Ruohan and Fei-Fei, Li and Liu, C. Karen},
  journal={arXiv preprint arXiv:2403.07788},
  year={2024}
}

@article{24arxiv_pi0,
  title={${\pi}_0$: A vision-language-action flow model for general robot control},
  author={Black, Kevin and Brown, Noah and Driess, Danny and Esmail, Adnan and Equi, Michael and Finn, Chelsea and Fusai, Niccolo and Groom, Lachy and Hausman, Karol and Ichter, Brian and others},
  journal={arXiv preprint arXiv:2410.24164},
  year={2024}
}

@article{25icra_dexmimicgen,
  title={DexMimicGen: Automated data generation for bimanual dexterous manipulation via imitation learning},
  author={Jiang, Zhenyu and Xie, Yuqi and Lin, Kevin and Xu, Zhenjia and Wan, Weikang and Mandlekar, Ajay and Fan, Linxi and Zhu, Yuke},
  journal={IEEE International Conference on Robotics and Automation},
  year={2025}
}

@inproceedings{25cvpr_maniptrans,
  title={{ManipTrans}: Efficient dexterous bimanual manipulation transfer via residual learning},
  author={Li, Kailin and Li, Puhao and Liu, Tengyu and Li, Yuyang and Huang, Siyuan},
  booktitle={IEEE/CVF Conference on Computer Vision and Pattern Recognition},
  pages={6991--7003},
  year={2025}
}

@inproceedings{25iclr_dextrack,
  title={{DexTrack}: Towards generalizable neural tracking control for dexterous manipulation from human references},
  author={Liu, Xueyi and Adalibieke, Jianibieke and Han, Qianwei and Qin, Yuzhe and Yi, Li},
  booktitle={International Conference on Learning Representations},
  year={2025}
}

@article{25arxiv_dexumi,
  title={{DexUMI}: Using human hand as the universal manipulation interface for dexterous manipulation},
  author={Xu, Mengda and Zhang, Han and Hou, Yifan and Xu, Zhenjia and Fan, Linxi and Veloso, Manuela and Song, Shuran},
  journal={arXiv preprint arXiv:2505.21864},
  year={2025}
}

@inproceedings{25rss_dex1b,
  title={{Dex1B}: Learning with 1B demonstrations for dexterous manipulation},
  author={Ye, Jianglong and Wang, Kuan and Yuan, Chengbo and Yang, Ruihan and Li, Yiming and Zhu, Jialiang and Qin, Yuzhe and Zou, Xueyan and Wang, Xiaolong},
  booktitle={Robotics: Science and Systems},
  year={2025}
}

@inproceedings{25corl_dexplore,
  title={{Dexplore}: Scalable neural control for dexterous manipulation from reference-scoped exploration},
  author={Xu, Sirui and Chao, Yu-Wei and Bian, Liuyu and Mousavian, Arsalan and Wang, Yu-Xiong and Gui, Liang-Yan and Yang, Wei},
  booktitle={Conference on Robot Learning},
  year={2025}
}

@article{26arxiv_unidex,
  title={{UniDex}: A robot foundation suite for universal dexterous hand control from egocentric human videos},
  author={Zhang, Gu and Xu, Qicheng and Zhang, Haozhe and Ma, Jianhan and He, Long and Bao, Yiming and Ping, Zeyu and Yuan, Zhecheng and Lu, Chenhao and Yuan, Chengbo and Liang, Tianhai and Tian, Xiaoyu and Shao, Maanping and Zhang, Feihong and Ding, Mingyu and Gao, Yang and Zhao, Hao and Zhao, Hang and Xu, Huazhe},
  journal={arXiv preprint arXiv:2603.22264},
  year={2026}
}

@inproceedings{24corl_opentelevision,
  title={Open-{TeleVision}: Teleoperation with immersive active visual feedback},
  author={Cheng, Xuxin and Li, Jialong and Yang, Shiqi and Yang, Ge and Wang, Xiaolong},
  booktitle={Conference on Robot Learning},
  year={2024}
}

@article{25arxiv_hermes,
  title={{HERMES}: Human-to-robot embodied learning from multi-source motion data for mobile dexterous manipulation},
  author={Yuan, Zhecheng and Wei, Tianming and Gu, Langzhe and Hua, Pu and Liang, Tianhai and Chen, Yuanpei and Xu, Huazhe},
  journal={arXiv preprint arXiv:2508.20085},
  year={2025}
}

@inproceedings{12humanoids_interactionmesh,
  title={Interaction mesh based motion adaptation for biped humanoid robots},
  author={Nakaoka, Shin'ichiro and Komura, Taku},
  booktitle={IEEE-RAS International Conference on Humanoid Robots},
  pages={625--631},
  year={2012},
  doi={10.1109/HUMANOIDS.2012.6651585}
}

@article{25arxiv_omniretarget,
  title={{OmniRetarget}: Interaction-preserving data generation for humanoid whole-body loco-manipulation and scene interaction},
  author={Yang, Lujie and Huang, Xiaoyu and Wu, Zhen and Kanazawa, Angjoo and Abbeel, Pieter and Sferrazza, Carmelo and Liu, C. Karen and Duan, Rocky and Shi, Guanya},
  journal={arXiv preprint arXiv:2509.26633},
  year={2025}
}

@article{25rss_demogen,
  title={Demogen: Synthetic demonstration generation for data-efficient visuomotor policy learning},
  author={Xue, Zhengrong and Deng, Shuying and Chen, Zhenyang and Wang, Yixuan and Yuan, Zhecheng and Xu, Huazhe},
  journal={arXiv preprint arXiv:2502.16932},
  year={2025}
}

@inproceedings{24cvpr_diff3f,
  title={Diffusion 3d features (diff3f) decorating untextured shapes with distilled semantic features},
  author={Dutt, Niladri Shekhar and Muralikrishnan, Sanjeev and Mitra, Niloy J},
  booktitle={2024 IEEE/CVF Conference on Computer Vision and Pattern Recognition (CVPR)},
  pages={4494--4504},
  year={2024},
  organization={IEEE}
}

@article{26arxiv_toporetarget,
  title={TopoRetarget: Interaction-Preserving Retargeting for Dexterous Manipulation},
  author={Wu, Jielin and Yao, Shenzhe and He, Guanqi and Liu, Xiaohan and Zeng, Zhaoqing and Jiang, Xiangrui and Yang, Han and Zhang, Wentao and Zhao, Hang},
  journal={arXiv preprint arXiv:2606.16272},
  year={2026}
}

@article{26arxiv_c2dex,
  title={C2Dex: Contact-Consistent Reconstruction and Retargeting for Dexterous Manipulation from Monocular Video},
  author={Ren, Jie and Jiang, Zhehao and Yang, Yinhong and Jia, Haorui and Jiang, Han and Li, Ben and Yao, Yao and Lin, Cheng and Shen, Qiu and Bing, Zhenshan and others},
  journal={arXiv preprint arXiv:2608.07045},
  year={2026}
}

@article{25arxiv_spider,
  title={Spider: Scalable physics-informed dexterous retargeting},
  author={Pan, Chaoyi and Wang, Changhao and Qi, Haozhi and Liu, Zixi and Bharadhwaj, Homanga and Sharma, Akash and Wu, Tingfan and Shi, Guanya and Malik, Jitendra and Hogan, Francois},
  journal={arXiv preprint arXiv:2511.09484},
  year={2025}
}

@article{26arxiv_doasido,
  title={Do as i do: Dexterous manipulation data from everyday human videos},
  author={Paliwal, Bhawna and Etukuru, Haritheja and Liang, William and Abbeel, Pieter and Shafiullah, Nur Muhammad Mahi and Malik, Jitendra},
  journal={arXiv preprint arXiv:2606.19333},
  year={2026}
}

@article{26arxiv_lingbotvla,
  title={A pragmatic vla foundation model},
  author={Wu, Wei and Lu, Fan and Wang, Yunnan and Yang, Shuai and Liu, Shi and Wang, Fangjing and Zhu, Qian and Sun, He and Wang, Yong and Ma, Shuailei and others},
  journal={arXiv preprint arXiv:2601.18692},
  year={2026}
}

@article{23tog_gs,
  title={3d gaussian splatting for real-time radiance field rendering.},
  author={Kerbl, Bernhard and Kopanas, Georgios and Leimk{\"u}hler, Thomas and Drettakis, George and others},
  journal={ACM Trans. Graph.},
  volume={42},
  number={4},
  pages={139--1},
  year={2023}
}

@article{26arxiv_regrind,
  title={A Minimalist Retargeting-Guided Reinforcement Learning Recipe for Dexterous Manipulation},
  author={Feng, Yunhai and Leung, Natalie and Wang, Jiaxuan and Yang, Lujie and Qi, Haozhi and Culbertson, Preston},
  journal={arXiv preprint arXiv:2607.11874},
  year={2026}
}

@article{26arxiv_deximit,
  title={Deximit: Learning bimanual dexterous manipulation from monocular human videos},
  author={Mu, Juncheng and Yang, Sizhe and Bao, Yiming and Bae, Hojin and Wei, Tianming and Xu, Linning and Li, Boyi and Xu, Huazhe and Pang, Jiangmiao},
  journal={arXiv preprint arXiv:2602.10105},
  year={2026}
}

@misc{wuji,
  title        = {{Wuji Technology}},
  howpublished = {\url{https://github.com/wuji-technology}}
}

@misc{franka,
  title        = {{Franka Robotics}},
  howpublished = {\url{https://github.com/frankarobotics}}
}

@software{mink,
  author = {Zakka, Kevin},
  title = {{Mink: Python inverse kinematics based on MuJoCo}},
  year = {2026},
  month = feb,
  version = {1.1.0},
  url = {https://github.com/kevinzakka/mink},
  license = {Apache-2.0}
}

@article{isaaclab,
  title={Isaac lab: A gpu-accelerated simulation framework for multi-modal robot learning},
  author={Mittal, Mayank and Roth, Pascal and Tigue, James and Richard, Antoine and Zhang, Octi and Du, Peter and Serrano-Munoz, Antonio and Yao, Xinjie and Zurbr{\"u}gg, Ren{\'e} and Rudin, Nikita and others},
  journal={arXiv preprint arXiv:2511.04831},
  year={2025}
}

@article{26nips_rtc,
  title={Real-time execution of action chunking flow policies},
  author={Black, Kevin and Galliker, Manuel and Levine, Sergey},
  journal={Advances in Neural Information Processing Systems},
  volume={38},
  pages={33383--33407},
  year={2026}
}
}

\end{document}